\documentclass[10pt,twocolumn,letterpaper]{article}

\usepackage{cvpr}      
\definecolor{pinkhighlight}{RGB}{255,182,193}
\definecolor{bluehighlight}{RGB}{0,0,255}
\definecolor{greenhighlight}{RGB}{100,180,100}
\usepackage{xcolor,colortbl}
\usepackage[ruled,vlined,linesnumbered]{algorithm2e}

\newcommand{\KC}[1]{{\color{black}{{ #1}}}}

\definecolor{cvprblue}{rgb}{0.21,0.49,0.74}
\usepackage[pagebackref,breaklinks,colorlinks,allcolors=cvprblue]{hyperref}

\def\paperID{1733} 
\def\confName{CVPR}
\def\confYear{2026}

\title{Your Classifier Can Do More: Towards Balancing the Gaps in Classification, Robustness, and Generation}

\author{
  \vspace{-20pt}\\
  {Kaichao Jiang$^{1,2,3}$, He Wang$^{4}$, Xiaoshuai Hao$^{5}$, Xiulong Yang$^6$, Ajian Liu$^{7}$, Qi Chu$^{8}$}\\ 
  {Yunfeng Diao$^{1,2,3}$\thanks{Corresponding authors: \,\href{mailto:diaoyunfeng@hfut.edu.cn}{\color{black}{diaoyunfeng@hfut.edu.cn}}}, Richang Hong$^{1}$}\\
  $^1$Hefei University of Technology, China
  $^2$Jianghuai Advance Technology Center, China\\
  $^3$Anhui Provincial Key Laboratory of Humanoid Robots, 
  China\\ $^4$AI Centre, University College London, UK 
  $^5$Xiaomi EV, China \\
  $^6$Central China Normal University, China \\
  $^7$Institute of Automation, Chinese Academy of Sciences, China  \\
  $^8$University of Science and Technology of China, China \\
  \texttt{\small 2022212327@mail.hfut.edu.cn, hewang@ucl.ac.uk, haoxiaoshuai@xiaomi.com,} \\ 
  \texttt{\small  yangxiulong@ccnu.edu.cn, ajianliu92@gmail.com, qchu@ustc.edu.cn}  \\
  \texttt{\small  diaoyunfeng@hfut.edu.cn, hongrc.hfut@gmail.com}  \\
}
\begin{document}
\maketitle
\begin{abstract}
Joint Energy-based Models (JEMs) are well known for their ability to unify classification and generation within a single framework. Despite their promising generative and discriminative performance, their robustness remains far inferior to adversarial training (AT), which, conversely, achieves strong robustness but sacrifices clean accuracy and lacks generative ability. This inherent trilemma—balancing classification accuracy, robustness, and generative capability—raises a fundamental question: \textit{Can a single model achieve all three simultaneously?} To answer this, we conduct a systematic energy landscape analysis of clean, adversarial, and generated samples across various JEM and AT variants. We observe that AT reduces the energy gap between clean and adversarial samples, while JEMs narrow the gap between clean and synthetic ones. This observation suggests a key insight: if the energy distributions of all three data types can be aligned, we might bridge their performance disparities. Building on this idea,  we propose Energy-based Joint Distribution Adversarial Training (EB-JDAT), a unified generative-discriminative-robust framework that maximizes the joint probability of clean and adversarial distribution. EB-JDAT introduces a novel min–max energy optimization to explicitly aligning energies across clean, adversarial, and generated samples. Extensive experiments on CIFAR-10, CIFAR-100, and ImageNet subsets demonstrate that EB-JDAT achieves state-of-the-art robustness while maintaining near-original accuracy and \KC{competitive generation quality of JEMs, effectively achieving a new trade-off frontier between accuracy, robustness, and generation.} The code is released at https://github.com/yujkc/EB-JDAT.

\end{abstract}    
\vspace{-0.5cm}
\section{Introduction}
\label{sec:intro}

\begin{figure}[t]
  \centering
  \includegraphics[width=\linewidth]{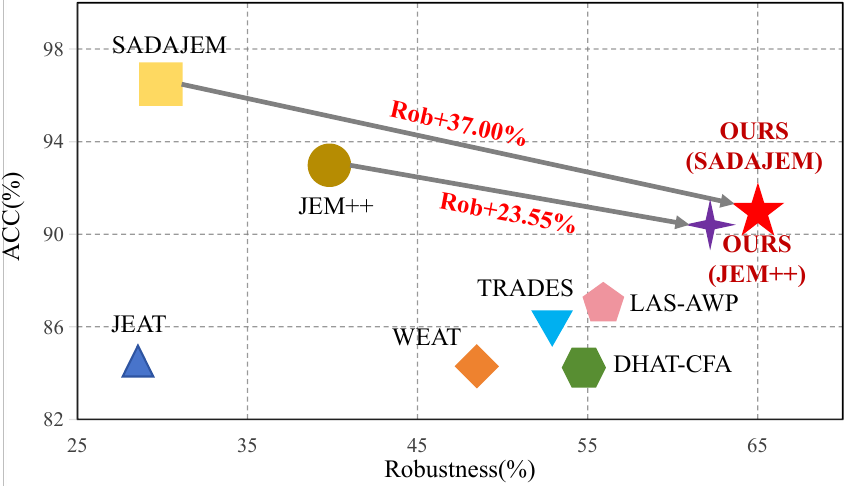} 
  \caption{Comparisons of SOTA AT-based methods on CIFAR-10 in terms of accuracy and robustness (AutoAttack). Our method achieves the best robustness while maintaining competitive standard accuracy.}
  \vspace{-0.6cm}
  \label{fig:robustness_jem}
\end{figure}

Recently, Joint Energy-based Models (JEMs) \cite{grathwohl2019your,yang2021jem++,yang2023towards} have garnered significant attention for their ability to integrate generative modeling within a discriminative framework. This approach effectively bridges classification and generation from an energy-based perspective and exhibits surprising inherent robustness. However, JEMs still exhibit limited robustness compared to Adversarial Training (AT)~\cite{mkadry2017towards}. On the other hand, AT~\cite{jia2022adversarial,hammoudeh2024provable,bortolussi2024robustness,zhang2025towards} is widely recognized as the most effective method for improving the robustness of discriminative classifiers. However, it typically sacrifices accuracy on clean data and lacks generative capability \cite{zhang2019theoretically}. Consequently, the inherent gap in classification, robustness and generation raises a natural question: Can a single model simultaneously achieve high classification accuracy, adversarial robustness, and generative performance? -- a goal that has been rarely explored.

\begin{figure*}
  \centering
  \includegraphics[width=\textwidth]{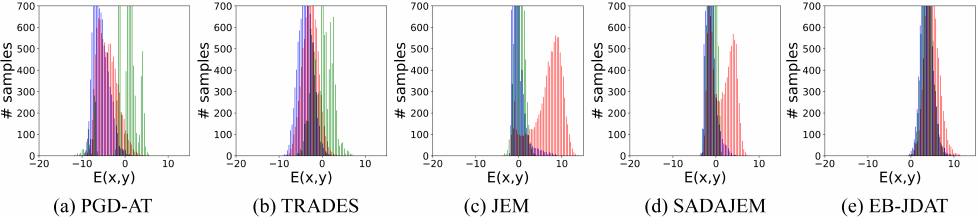} 
  \caption{Distributions of the $E_\theta(\textbf{x},y)$ of adversarial samples for PGD-20 vs. generated samples vs. clean samples on CIFAR-10. \colorbox{greenhighlight}{\;}indicates generated samples, \colorbox{pinkhighlight}{\;}indicates adversarial samples, \colorbox{bluehighlight}{\;}indicates clean samples.}
  \vspace{-0.4cm}
  \label{fig:robustness_vs_gen}
\end{figure*}

To answer this question, we investigate the underlying causes of the performance gap between AT~\cite{mkadry2017towards,zhang2019theoretically} and JEMs~\cite{grathwohl2019your,yang2023towards} in terms of accuracy, generative capability, and adversarial robustness from an energy perspective. We compute the energy distributions of clean, adversarial, and generated samples, with results shown in~\cref{fig:robustness_vs_gen}. Further, we compute the mean and variance of the one-to-one energy differences between clean and adversarial samples, as reported in~\cref{table:robustness_onetoone}. The energy distributions shows that AT reduces the energy gap between clean and adversarial samples, thereby bringing robustness. In contrast, JEMs reduce the gap between clean and generated samples, which brings generative capability and higher accuracy. Based on these insights, we draw a key conclusion: \textit{if the energy distributions of three data types can be aligned, we might unify the strengths of AT and JEMs, resolving the inherent trilemma}.
\begin{table}[t]
    \caption{Mean and variance of one‐to‐one energy differences between clean and adv. Results in bold indicate the best.}
    \label{table:robustness_onetoone}
    \small
    \centering
    \setlength{\tabcolsep}{13pt}
    \begin{tabular}{@{}lcccc@{}}
    \toprule
    Model & Mean$\downarrow$& Variance$\downarrow$ & PGD(\%)$\uparrow$ \\ 
    \midrule
    Standard & 10.18 & 3.27 & 0.06 \\
    PGD-AT \cite{mkadry2017towards}   & 1.46  & 0.83 & 55.08 \\
    TRADES \cite{zhang2019theoretically}  & 1.01  & 0.45 & 56.10 \\
    JEM \cite{grathwohl2019your}     & 5.93  & 7.30 & 7.28  \\
    SADAJEM \cite{yang2023towards}  & 2.66  & 3.83 & 43.64 \\
    \midrule
    EB-JDAT & \textbf{0.89} & \textbf{0.31} & \textbf{66.12} \\
    \bottomrule
  \end{tabular}
  \vspace{-0.3cm}
  \end{table}

\vspace{-0.4cm}
Inspired by this idea, we align the energy distributions of clean, adversarial and generated data through a joint energy-based model of $p_{\theta}(\mathbf{x}, \tilde{\mathbf{x}}, y)$, with $\mathbf{x}$ clean samples, $\tilde{\mathbf{x}}$ adversarial examples, $y$ class labels and $\theta$ the model parameters. Through Bayesian decomposition, $p_{\theta}(\mathbf{x}, \tilde{\mathbf{x}}, y)$ can be further factorized as the data distribution $p_{\theta}(\mathbf{x})$, adversarial distribution $p_{\theta}(\tilde{\mathbf{x}} \mid \mathbf{x})$ and $p_{\theta}(y \mid \tilde{\mathbf{x}}, \mathbf{x})$. While $p_{\theta}(\mathbf{x})$ can be computed via approximated sampling methods~\cite{nijkamp2020anatomy} and $p_{\theta}(y \mid \tilde{\mathbf{x}}, \mathbf{x})$ is simply a cross-entropy objective for robust classification, computing $p_{\theta}(\tilde{\mathbf{x}} \mid \mathbf{x})$ is not straightforward because the full adversarial distribution is not observed during the training phase. However, our key observation is that adversarial perturbations almost surely move samples out of the original data manifold, which lies in high-density areas, into low-density areas where the classifier is more likely to fail, as illustrated by the weakly robust classifiers JEM \cite{grathwohl2019your} and SADAJEM \cite{yang2023towards} in \cref{fig:robustness_vs_gen}. Motivated by this observation, we propose a new min-max energy optimization to approximate $p_{\theta}(\tilde{\mathbf{x}} \mid \mathbf{x})$. During the sampling phase, we search for high-energy adversarial examples, then minimize the energy gap between adversarial and clean samples during the training phase, thereby pulling the adversarial samples back into low-energy regions. We name our proposed method as Energy-based Joint Distribution Adversarial Training, or EB-JDAT. Extensive experiments across multiple datasets and attacks show that EB-JDAT achieves 68.76\%, 35.63\% and 32.40\% robustness (under AutoAttack \cite{croce2020reliable}) on CIFAR-10, CIFAR-100 and ImageNet subset, outperforming the SOTA AT methods by +10.78\%, +4.70\% and +7.88\% respectively, while maintaining near original clean accuracy and reasonable  generative capability, which is competitive to most advanced JEMs.

\section{Related Work}
\label{sec:rel}
\vspace{-0.2cm}
\paragraph{Adversarial Defense.}
Since the discovery of adversarial vulnerabilities in deep learning models \cite{yuan2021automa,ma2023transferable,yang2023revisiting, bao2025aucpro,qin2025mixbridge,sun2025diffmark,hao2025mimo}, numerous defense strategies have been proposed to mitigate this security challenge. Prominent approaches include input denoising \cite{li2022feature}, gradient regularization \cite{li2023revisiting,jia2022prior}, defensive distillation \cite{huang2023boosting}, and adversarial training (AT) \cite{mkadry2017towards,fang20253sat,zhang2025towards}. Recent studies confirm that AT remains among the most effective defense methods \cite{sengphanith2023evaluating}, significantly enhancing model robustness by incorporating adversarial examples into the training process. Notably, \cite{mkadry2017towards} first formulated AT as a saddle-point optimization problem (PGD), laying the foundational framework for subsequent robust training methodologies. As adversarial attack methods have evolved, various novel strategies have emerged, refining conventional AT methods \cite{jia2022adversarial,hammoudeh2024provable,bortolussi2024robustness,fang20253sat,zhang2025towards}. However, many of these strategies prioritize robustness at the expense of clean accuracy. which is the principal evaluation metric in visual classification, highlighting an inherent trade-off in AT. To mitigate this issue, \cite{zhang2019theoretically} proposed TRADES, which constrains the divergence in output probability between clean and adversarial examples. Additionally, \cite{co2022jacobian} combined Jacobian regularization with model ensembling, introducing the Jacobians ensemble approach that effectively balances robustness and accuracy against universal adversarial perturbations. Furthermore, DUCAT \cite{wang2024new} assigns a dummy class to each original class to absorb hard adversarial examples during training and remaps them back at inference to mitigating the trade-off. Although these approaches represent notable advancements, they still suffer from substantial accuracy degradation relative to standard classifiers, while offering only modest enhancements in robustness. Therefore, achieving a trade-off between accuracy and robustness remains an unresolved challenge in AT.
\vspace{-0.5cm}
\paragraph{Robust Energy-based Model.}
Recent studies have indicated that features extracted by robust classifiers are more closely aligned with human perceptual mechanisms compared to those derived from standard classifiers \cite{wangunified, deng2025robust}. The Joint Energy-based Model (JEM) \cite{grathwohl2019your} reformulates conventional softmax classifiers into an energy-based model (EBM), facilitating hybrid discriminative-generative modeling and clarifying the connection between robustness and generative modeling. More recent efforts have focused on improving training efficiency and speed in JEM \cite{yang2021jem++,yang2023towards}, effectively bridging the gap between classification and generation. Additionally, several approaches specifically focus on enhancing robustness, from an energy perspective. JEAT \cite{zhu2021towards} initially demonstrated a link between AT and EBM, highlighting that despite differences in how these methods modify the energy function, they ultimately employ comparable contrastive strategies. Subsequently, \cite{mirza2024shedding} reinterpreted robust classifiers within an energy-based framework and analyzed the energy landscape to reveal differential effects of targeted and non-targeted attacks during AT. \KC{Methodologically, previous work \cite{mirza2024shedding,lee2020adversarial,korst2022adversarial,yin2022learning} typically treats AT as an external regularization for JEMs by optimizing a robust classification objective $ p_{\theta}(y|\tilde{{\mathbf{x}}})$). JEAT \cite{zhu2021towards} injects adversarial samples directly into JEM training to model ${p}_{\theta}(\tilde{\mathbf{x}}, y)$, but neglects the intrinsic realationship between clean and adversarial data. In contrast, we model the full joint distribution $p_{\theta}(\mathbf{x}, \tilde{\mathbf{x}}, y)$, explicitly narrowing the energy gap among clean, adversarial, and generated samples, thereby more accurately capturing both clean and adversarial data distributions.}

\vspace{-0.2cm}
\section{Methodology}
\subsection{Preliminaries}
\paragraph{Energy-based Models (EBMs).}
From an energy perspective, the clean data distribution ${p}_{\theta}(\mathbf{x})$ can be explicitly parameterized through an energy function \cite{grathwohl2019your}:
\begin{equation}\label{Eq2}
p_{\theta}(\mathbf{x})=\frac{\exp \left(-E_{\theta}(\mathbf{x})\right)}{Z(\theta)}=\frac{\sum_{y \in \mathbf{y}} \exp \left(f_{\theta}(\mathbf{x})[y]\right)}{Z(\theta)},
\end{equation}
where $E_{\theta}(\cdot):\mathbb{R}^{D} \rightarrow \mathbb{R}$ denotes the energy function, $f_\theta (x)[y]$ denotes the $y^{\mathrm{th}}$ index of $f_\theta [y]$, i.e., the logit corresponding the $y^{\mathrm{th}}$ class label. This energy function assigns each data point a scalar value representing its density within the continuous data distribution. The normalizing constant $Z(\theta)=\int_{\mathbf{x}} \exp(-E_{\theta}(\mathbf{x}))$ ensures the proper normalization of probabilities. During training, the energy function $E_{\theta}(\mathbf{x})$ is optimized to assign lower energy values to high-density regions (e.g., training data) and higher energy values to low-density regions (e.g., adversarial samples). To optimize the energy-based model, a common strategy is to perform maximum likelihood estimation of the parameters $\theta$, in which the gradient of the log-likelihood is expressed as follows:
\begin{equation}
\label{Eq3}
\frac{\log p_{\theta}(\mathbf{x})}{\partial \theta} = \mathbb{E}_{p_{\theta}(\mathbf{x'})} \left[ \frac{\partial E_{\theta}(\mathbf{x'})}{\partial \theta} \right] - \frac{\partial E_{\theta}(\mathbf{x})}{\partial \theta},
\end{equation}
where $\mathbf{x'}$ is simply a sample obtained by sampling the distribution of models.
\vspace{-0.3cm}
\paragraph{Joint Energy-based Models (JEMs)}
JEM \cite{grathwohl2019your} reinterprets the standard softmax classifier within the framework of an EBM. Specifically, it redefines the logits of the classifier $f_\theta$ to model the joint distribution $p_{\theta}(\textbf{x},y)$ of the input data $\textbf{x}$ and their labels $y$ using an energy-based approach:
\begin{equation}
\label{jem}
p_{\theta}(\mathbf{x}, y)=\frac{\exp \left(f_{\theta}(\mathbf{x})[y]\right)}{Z(\theta)}.
\end{equation}

In JEMs \cite{grathwohl2019your,yang2021jem++,yang2023towards}, the model parameters $\theta$ are optimized by maximizing $\log p_{\theta}(\mathbf{x},y)$:
\begin{equation}
\label{pxy}
\log p_{{\theta}}(\mathbf{x}, y)=\log p_{{\theta}}(y \mid \mathbf{x})+\log p_{{\theta}}(\mathbf{x}).
\end{equation}
\paragraph{Adversarial Training (AT).}
Adversarial training can be viewed as a min-max optimization problem that integrates inner loss maximization and outer loss minimization \cite{mkadry2017towards}. Given data $\mathbf{x} \in \mathbf{X}$, its corresponding label $y \in \mathbf{y}$, and a classification model $f_{\theta}$, the objective of the inner maximization is to identify the most aggressive adversarial examples by maximizing the loss. Subsequently, the outer minimization seeks to minimize the average loss of these adversarial examples. The specific optimization process is detailed in:
\begin{equation}\label{adv}
\min _{\theta} \mathbb{E}_{(\mathbf{x}, {y}) \sim \mathcal{D}}\left[\max _{\|\tilde{\mathbf{x}}-\mathbf{x}\| \in \Omega} \mathbf{L}(\tilde{\mathbf{x}}, {y} ; \theta)\right],
\end{equation}
where $\mathcal{D}$ denotes the training data, $\mathbf{L}$ represents the loss function (typically cross-entropy for classification) , $\Tilde{\mathbf{x}}$ is the adversarial example corresponding to $\mathbf{x}$, and $\Omega$ is the perturbation space.

\subsection{Motivation}
AT\cite{mkadry2017towards} has demonstrated strong robustness against adversarial attacks. However, the adversarial samples introduced during training lie outside the true data manifold, causing the model to overemphasize these adversarial features while neglecting meaningful patterns in clean data \cite{ilyas2019adversarial}. Consequently, AT-trained models exhibit a noticeable accuracy gap compared to standard classifiers. In contrast, JEMs have shown the ability to reinterpret discriminative classifiers as generative models~\cite{grathwohl2019your}, achieving notable performance in both image classification and generation. While JEMs exhibit improved adversarial robustness relative to standard models, a significant robustness gap still remains when compared to AT-trained methods. This naturally raises a new research question: \textit{Can we combine the strengths of AT and JEMs to develop a unified framework that mitigates, or even eliminates, triple trade-off between accuracy, robustness and generative performance?} To answer this question, we begin by visualizing the energy distributions of clean vs. adversarial vs. generated samples to investigate the source of the performance gains in JEMs and AT, which are shown in \cref{fig:robustness_vs_gen}. 
\vspace{-0.3cm}
\paragraph{Analysis of the Robustness.}

We first analyze the energy distributions of clean vs. adversarial samples. Obviously, for models trained by AT, the energy distributions of clean and adversarial samples almost completely overlap. While for JEMs, the two distributions partially overlap but less than that observed in AT-based methods. Further, We compute the mean and variance of one-to-one energy differences between clean samples and their corresponding adversarial examples and report the results in \cref{table:robustness_onetoone}. By analyzing these results, we have two findings: 1) the greater the overlap between the energy distributions of clean and adversarial samples, the higher the model's robustness. 2) JEMs implicitly reduce the gap between adversarial and clean data, although the distribution of adversarial samples is not explicitly incorporated during training. This also explains the source of robustness in JEMs.
\vspace{-0.3cm}
\paragraph{Analysis of the Generative Capability.}

Next, we investigate the classification and generative gaps by analyzing the energy distributions of real vs. generated samples in \cref{fig:robustness_vs_gen}. Compared to AT, JEMs exhibit a significantly greater overlap between the generated and real data distributions. This closer alignment indicates that JEMs more accurately capture the true data distribution, thereby enhancing both generative quality and classification accuracy. Based on these empirical evidences, we derive an key observation: a generative classifier that explicitly minimizes the energy divergence among clean, generated, and adversarial data distribution can effectively bridge the gaps in classification, robustness, and generative capability.

\subsection{Energy-based Joint Distribution Adversarial Training}
To explicitly minimize the energy divergence among clean, generated, and adversarial data distribution, a straightforward idea is to jointly model the clean data distribution, the adversarial distribution and the classifier via learning the joint probability $p_{\theta}(\mathbf{x}, \tilde{\mathbf{x}}, y)$:
\begin{equation}\label{Eq1}
\begin{split}
p_{\theta}(\mathbf{x}, \tilde{\mathbf{x}}, y)
&= p_{\theta}(y \mid \tilde{\mathbf{x}}, \mathbf{x})\, p_{\theta}(\tilde{\mathbf{x}}, \mathbf{x})\\
&= p_{\theta}(y \mid \tilde{\mathbf{x}}, \mathbf{x})\, p_{\theta}(\tilde{\mathbf{x}} \mid \mathbf{x})\, p_{\theta}(\mathbf{x}) ,
\end{split}
\end{equation} 
where $\theta$ denotes the model parameters. The conditional distribution ${p}_{\theta}\left({y} | \tilde{\mathbf{x}}, {\mathbf{x}}\right)$ denotes cross-entropy objective for robust classification, such as AT~\cite{mkadry2017towards}. $p_{\theta}(\tilde{\mathbf{x}}, \mathbf{x})$ aims to capture the joint distribution of clean and adversarial examples, and it can be further decomposed using Bayesian theorem as $p_{\theta}(\tilde{\mathbf{x}} \mid \mathbf{x}) p_{\theta}(\mathbf{x})$. Similar to \cref{Eq2}, the adversarial distribution $p_{\theta}(\tilde{\mathbf{x}} \mid \mathbf{x})$ can also be parameterized through an energy-based formulation:

\begin{equation}\label{eq:energy_adv}
p_{\theta}(\tilde{\mathbf{x}}|\mathbf{x})=\frac{\exp \left(-E_{\theta}(\tilde{\mathbf{x}}| \mathbf{x})\right)}{\tilde{Z}_{\theta}}=\frac{\sum_{y \in \mathbf{y}} \exp \left(f_{\theta}(\tilde{\mathbf{x}}|\mathbf{x})[y]\right)}{\tilde{Z}_{\theta}},
\end{equation}
where $\tilde{Z}(\theta)=\int_{\tilde{\mathbf{x}}} \exp(-E_{\theta}(\tilde{\mathbf{x}}|\mathbf{x}))$. A natural choice to optimize the model parameters $\theta$ is to maximize the log-likelihood of $p_{\theta}(\mathbf{x}, \tilde{\mathbf{x}}, y)$:
\begin{align}\label{eq:likelihood}
\log p_{\theta}(\mathbf{x}, \tilde{\mathbf{x}}, y)
&= \log p_{\theta}(y \mid \tilde{\mathbf{x}}, \mathbf{x}) \notag\\
&\quad + \log p_{\theta}(\tilde{\mathbf{x}} \mid \mathbf{x})
+ \log p_{\theta}(\mathbf{x}) ,
\end{align}
where $\log p_{\theta}(y\mid\tilde{\mathbf{x}}, \mathbf{x})$ is simply a cross-entropy objective for robust classifier, $\log p_{\theta}(\mathbf{x})$ can be estimated via various sampling methods \cite{welling2011bayesian}. In our method, \cref{Eq3} is approximated by \cite{nijkamp2020anatomy}:
\begin{equation}
\label{Eq4}
\frac{\partial {\log} p_{\theta}(\mathbf{x})}{\partial \theta} \approx \frac{\partial}{\partial \theta}\left[\frac{1}{L_{1}} \sum_{i=1}^{L_{1}} E_{\theta}\left(\mathbf{x}_{i}^{+}\right)\right. \\\left.-\frac{1}{L_{2}} \sum_{i=1}^{L_{2}} E_{\theta}\left(\mathbf{x}_{i}^{-}\right)\right] ,
\end{equation}
where $\left\{\mathbf{x}_{i}^{+}\right\}_{i=1}^{L_{1}}$ denote all the training samples in a batch, and $\left\{\mathbf{x}_{i}^{-}\right\}_{i=1}^{L_{2}}$ are i.i.d samples drawn from $p_{\theta}(\mathbf{x})$ via Stochastic Gradient Langevin Dynamics (SGLD) \cite{welling2011bayesian}:
\begin{equation}
\label{SGLD}
\mathbf{x}_{t+1}^{-}=\mathbf{x}_{t}^{-}+\frac{c^{2}}{2} \frac{\partial \log p_{\theta}\left(\mathbf{x}_{t}^{-}\right)}{\partial \mathbf{x}_{t}^{-}}+c \epsilon, c >0 ,
\epsilon \in \mathbf{N}(0, \mathbf{I}) ,
\end{equation}
where $c$ is a step size, $\mathbf{N}$ denotes the normal distribution, and $\mathbf{I}$ is the identity matrix. Therefore, the key to optimizing \cref{eq:likelihood} is to estimate $\log\ p_{\theta}(\tilde{\mathbf{x}} \mid \mathbf{x})$. Although the full adversarial distribution is not observed during training, we start from a straightforward yet key observation from an energy view: adversarial samples typically appear near the boundaries or even outside the support of the original data distribution, situating them in low-density, thus high-energy regions. Additionally, by visualizing the energy distributions of clean and adversarial samples in \cref{fig:robustness_vs_gen} and \cref{table:robustness_onetoone}, we observe that, under robust classifiers, these distributions are nearly indistinguishable. This suggests that, after AT, adversarial examples are pulled back from high-energy area into low-energy (i.e., high-density) areas, close to clean samples. This inspires us to propose a min-max optimization framework to compute $\log p_{\theta}(\tilde{\mathbf{x}} \mid \mathbf{x})$:
\begin{equation}\label{eq:min_max}
\min _{\theta} \mathbb{E}_{(\mathbf{x}, y) \sim \mathcal{D}}\left[\max _{\|\tilde{\mathbf{x}}-\mathbf{x}\| \in \Omega}\left(E_{\theta}(\tilde{\mathbf{x}} \mid \mathbf{x})-E_{\theta}(\mathbf{x})\right)\right] ,
\end{equation}
where $\mathcal{D}$ denotes training data. The inner maximization problem aims to move the adversarial examples out of the high-density area, which corresponds to minimize the joint probability $\log {p_{\theta}}\left((\tilde{\mathbf{x}} \mid \mathbf{x}), y\right)$. 
Specifically, we update SGLD process along the opposite direction of $\nabla_{\mathbf{x}}\log {p_{\theta}}\left((\tilde{\mathbf{x}} \mid \mathbf{x}), y\right)$ to sample the adversarial examples:

\begin{equation}\label{SGLD_adv}
\tilde{\mathbf{x}}_{t+1}=\tilde{\mathbf{x}}_{t}-\frac{c^{2}}{2} \frac{\partial\log {p_{\theta}}\left((\tilde{\mathbf{x}} \mid \mathbf{x}), y\right)}{\partial \tilde{\mathbf{x}}_{t}},
\end{equation}
On the other hand, the goal of the outer minimization problem is to find model parameters so that the energy difference between adversarial and clean samples is minimized, i.e. pulling back the adversarial examples to the high-density area. Unlike existing robust classifiers only maximize the conditional probability $\log p_{\theta}(y\mid\tilde{\mathbf{x}})$, we capture the full adversarial data distribution during training. Similar for \cref{Eq4}, $\frac{\partial {\log} p_{\theta}(\tilde{\mathbf{x}}|\mathbf{x})}{\partial \theta}$ can be approximated as:

\begin{equation}\label{sample_adv}
\frac{\partial {\log} p_{\theta}(\tilde{\mathbf{x}}|\mathbf{x})}{\partial \theta} \approx \frac{\partial}{\partial \theta}\left[\frac{1}{L_{1}} \sum_{i=1}^{L_{1}} E_{\theta}\left(\mathbf{x}_{i}^{+}\right)-\frac{1}{L_{2}} \sum_{i=1}^{L_{2}} E_{\theta}\left(\tilde{\mathbf{x}}_{i}|\mathbf{x}_{i}^{+}\right)\right].
\end{equation}
\KC{Thus, EB-JDAT constitutes a genuine conditional EBM for ${p}_{\theta}(\tilde{{\mathbf{x}}}| {\mathbf{x}})$.} Our full Algorithm is presented in \cref{alg:EB-JDAT}.

\begin{algorithm}[t]
\caption{Energy-based Joint Distribution AT}
\label{alg:EB-JDAT}
\SetAlgoLined
\textbf{Input}: $\mathbf{x}$: training data; $N_{tra}$: the number of training iterations; $M$: sampling iterations; $M_{adv}$: adversarial sampling iterations; $M_\theta$: sampling iterations for $\theta$ \;
\textbf{Output}: $\theta$: network weight\;
\textbf{Init}: randomly initialize $\theta$\;

\For{$i = 1$ to $N_{tra}$}{
    Sample a mini-batch $\{\mathbf{x},y\}_i$\;
    Obtain ${\mathbf{x}}_0$ via random noise\;
    \For{$m = 1$ to $M$}{
    Sample ${\mathbf{x}}_t$ via \cref{SGLD}\;
    }
    Compute $h_1 = \frac{\partial \log p_{\theta}({\mathbf{x}})} {\partial\theta}$ via \cref{Eq4}\;
    \textbf{//MIN-MAX Optimization for $p_{\theta}(\tilde{{\mathbf{x}}} | {\mathbf{x}})$} \\
    Obtain $\tilde{\mathbf{x}}_0$ from ${\mathbf{x}}$ with a perturbation\;
    \For{$m_{adv} = 1$ to $M_{adv}$}{
    Sample $\tilde{{\mathbf{x}}}_t$ via \cref{SGLD_adv}\;
    }
    Compute $h_2 = \frac{\partial \log p_{\theta}(\tilde{{\mathbf{x}}} | {\mathbf{x}})}{\partial \theta}$ via from \cref{sample_adv}\;
    Compute $h_3 = \frac{\partial \log p_{\theta}(y | {\mathbf{x}}, \tilde{{\mathbf{x}}})}{\partial \theta}$ via cross-entropy loss\;
    ${h}_{\theta} = h_1 + h_2 + h_3$\;
    \For{t = 1 to $M_\theta$}{
        update $\theta$\  via ${h}_{\theta}$;

        }
    }
    
\Return $\theta$\;

\end{algorithm}

\section{Experiments}
\label{Experiments}
\subsection{Experimental Setup}
\label{Experimental Setup}
We conduct analysis and comparison experiments on several standard classifier datasets, including CIFAR-10 \cite{krizhevsky2009learning}, CIFAR-100 \cite{krizhevsky2009learning} and ImageNet \cite{deng2009imagenet}. Given that most AT methods utilize WideResNet \cite{zagoruyko2016wide} architectures, we select the commonly used WRN28-10 as the baseline network. All experiments were carried out on 3090 GPU. During the entire training process, we follow the default settings and SGLD hyperparameters of JEMs \cite{yang2021jem++,yang2023towards}, default seed is 1, and the learning rate is set to 0.01. The initialization perturbation for adversarial samples is set to 8/255, and adversarial sampling steps is 5, while the perturbations are constrained using an $\ell_\infty$ norm. Note that EB-JDAT is a general optimization framework for JEM that is compatible with various JEM-based models \cite{grathwohl2019your, yang2021jem++,yang2023towards,yang2023m}. Therefore, we integrate EB-JDAT with SADAJEM \cite{yang2023towards} and JEM++ \cite{yang2021jem++}, two faster and more stable JEMs, to achieve improved performance.

\begin{table*}[ht] 
\centering  
\caption{Comparative robustness (\%) analysis with different AT methods. Results in bold indicate the best. Across 3 runs with random seeds from \{1,2,3\}.}

\label{table:robustness_comparison}
\small
\tabcolsep 12pt
\begin{tabular*}{\textwidth}{cccc||ccc}
\toprule
& \multicolumn{3}{c||}{CIFAR-10} & \multicolumn{3}{c}{CIFAR-100} \\ 
Method & Clean & PGD-20 & AA & Clean & PGD-20 & AA \\ 
\midrule
Standard Training& 96.10 & 0.06 & 0 & 80.73 & 0 & 0\\
MART \cite{wang2019improving} & 82.99 & 55.48 & 50.67 & 54.69 & 31.90 & 27.25 \\ 
AWP \cite{wu2020adversarial} & 82.67 & 57.21 & 51.90 & 57.94 & 33.75 & 28.90 \\ 
LBGAT \cite{cui2021learnable} & 86.22 & 52.66 & 50.23 & 58.64 & 32.75 & 28.33 \\ 
LAS-AWP \cite{jia2022adversarial}& 87.74 & 60.16 & 55.52 & 64.89 & 36.36 & 30.77 \\
UIAT \cite{dong2023enemy} & 82.94 & 58.12 & 52.17 & 57.65 & 33.91 & 29.03 \\
SGLR \cite{li2024soften} & 85.72 & 56.10 & 53.40 & 61.02 & 32.98 & 28.50 \\ 
TDAT \cite{tong2024taxonomy} & 82.25 & 56.03 & 54.06 & 57.32 & 33.17 & 26.61 \\ 
AGR-TRADES \cite{tong2024balancing} & 86.50 & 52.56 & 52.56 & 58.17 & 28.34 & 25.84 \\ 
DHAT-CFA \cite{zhang2025towards} & 84.49 & 62.38 & 54.05 & 61.54 & 37.15 & 30.93 \\ 
\midrule
EB-JDAT-JEM++ & 90.30$\pm$0.35  & 64.88$\pm$0.21 & 64.78$\pm$0.23 & 68.05$\pm$0.46 & 37.21$\pm$0.35 & 35.19$\pm$0.29 \\ 
EB-JDAT-SADAJEM & \textbf{90.37$\pm$0.21} & \textbf{68.76$\pm$0.19} & \textbf{66.12$\pm$0.23} & \textbf{68.32$\pm$0.32} & \textbf{38.42$\pm$0.26} & \textbf{35.57$\pm$0.24} \\ 
\bottomrule
\end{tabular*}
\vspace{-0.1cm}
\end{table*}

\begin{table*}[ht]
\caption{Camparative robustness (\%) analysis with AT methods that utilize generative models for data augmentation. Time is 3090 GPU hours. Results in bold indicate the best.}
\label{table:robustness_comparison_extradata}
\small
\tabcolsep 20pt 
\begin{tabular*}{\textwidth}{c c c c c c}
\toprule
Data   & Generate data &Epoch & Clean & AA & Time \\
\midrule
\multicolumn{5}{l}{\textbf{CIFAR-10}} \\
SCORE \cite{pang2022robustness}     &1M  &400  & 88.10 & 61.51 & $\approx$ 1,438 h\\
The method proposed by \cite{gowal2021improving}   &100M  &2000  & 87.50 & 63.38 & $\approx$ 719,460 h\\
Better DM \cite{wang2023better}     &1M  &400  & \textbf{91.12} & 63.35 & $\approx$ 1,438 h\\
{EB-JDAT-JEM++  }  &n/a  &100  &{{90.37}}  & \textbf{64.61} & \textbf{31.66 h}\\
{EB-JDAT-SADAJEM  }  &n/a  &100  &{{90.39}}  & \textbf{66.30} & \textbf{66.64 h}\\
\midrule
\multicolumn{5}{l}{\textbf{CIFAR-100}} \\
SCORE \cite{pang2022robustness}     &1M  &400  & 62.08 & 31.40 & $\approx$ 1,438 h\\
Better DM \cite{wang2023better}     &1M  &400  & 68.06 & \textbf{35.65} & $\approx$ 1,438 h\\
{EB-JDAT-JEM++  } &n/a  &100  & {{68.09}}  & {{35.21}} &\textbf{31.66 h} \\
{EB-JDAT-SADAJEM  } &n/a  &100  & \textbf{{68.35}}  & {{35.63}} &\textbf{66.70 h}\\
\bottomrule
\end{tabular*}
\vspace{-0.5cm}
\end{table*}

\begin{table}[t]
  \centering
  \caption{Comparison on ImageNet-Subset in terms of accuracy and robustness. Results in bold indicate the best.}
  \label{table:imagenet_subset_acc_robust}
  \small
  \setlength{\tabcolsep}{10pt}
  \begin{tabular}{lccc}
    \toprule
    Method & Acc(\%)$\uparrow$ & PGD(\%)$\uparrow$ & AA(\%)$\uparrow$ \\
    \midrule
    MART \cite{wang2019improving}  & 49.30 & 21.84 & 17.40 \\
    LAS-AT \cite{jia2022adversarial}& 50.66 & 27.34 & 21.78 \\
    WEAT \cite{mirza2024shedding}& 61.90 & 30.27 & 24.52 \\
    \midrule
    EB-JDAT-JEM++  & \textbf{63.02} & \textbf{34.50} & \textbf{32.40} \\
    \bottomrule
  \end{tabular}
  \vspace{-0.4cm}
\end{table}

\begin{table}[t]
  \centering
  \caption{Comparison of different hybrid models on CIFAR-10, IF (informative initialization). Results in bold indicate the best.}
  \label{table:performance_comparison_gen}
  \small
  \setlength{\tabcolsep}{5.5pt}           
  \begin{tabular}{ccccc}
    \toprule
    Model & Acc(\%)$\uparrow$  & AA(\%)$\uparrow$ & FID$\downarrow$ & IS$\uparrow$ \\
    \midrule
    \multicolumn{4}{l}{\textbf{JEM variants}}\\
    JEM \cite{grathwohl2019your} & 92.90 & 4.28 & 38.40 & 8.76 \\
    JEM++ \cite{yang2021jem++}   & 93.73 & 41.06 & 37.12 & 8.29 \\
    SADAJEM \cite{yang2023towards} & \textbf{96.03} &29.63 & \textbf{17.38} & 8.07 \\
    \multicolumn{4}{l}{\textbf{Energy-based AT}}\\
    JEAT \cite{zhu2021towards}   & 85.16 & 28.43 & 38.24 & 8.80 \\
    WEAT \cite{mirza2024shedding}(PCA) & 83.36 &49.02 & 30.74 & \textbf{8.97} \\
    WEAT \cite{mirza2024shedding}(IF)  & 83.36 &49.02 & 177.92 & 3.50 \\
    \midrule
    EB-JDAT-JEM++   & 90.37 & \textbf{64.61} & 39.67 & 7.66 \\
    EB-JDAT-SADAJEM & 90.39 & \textbf{66.30} & 27.42 & 8.05 \\
    \bottomrule
  \end{tabular}
  \vspace{-0.5cm}
\end{table}

\begin{figure*}[!h]
  \centering
  \includegraphics[width=0.9\linewidth]{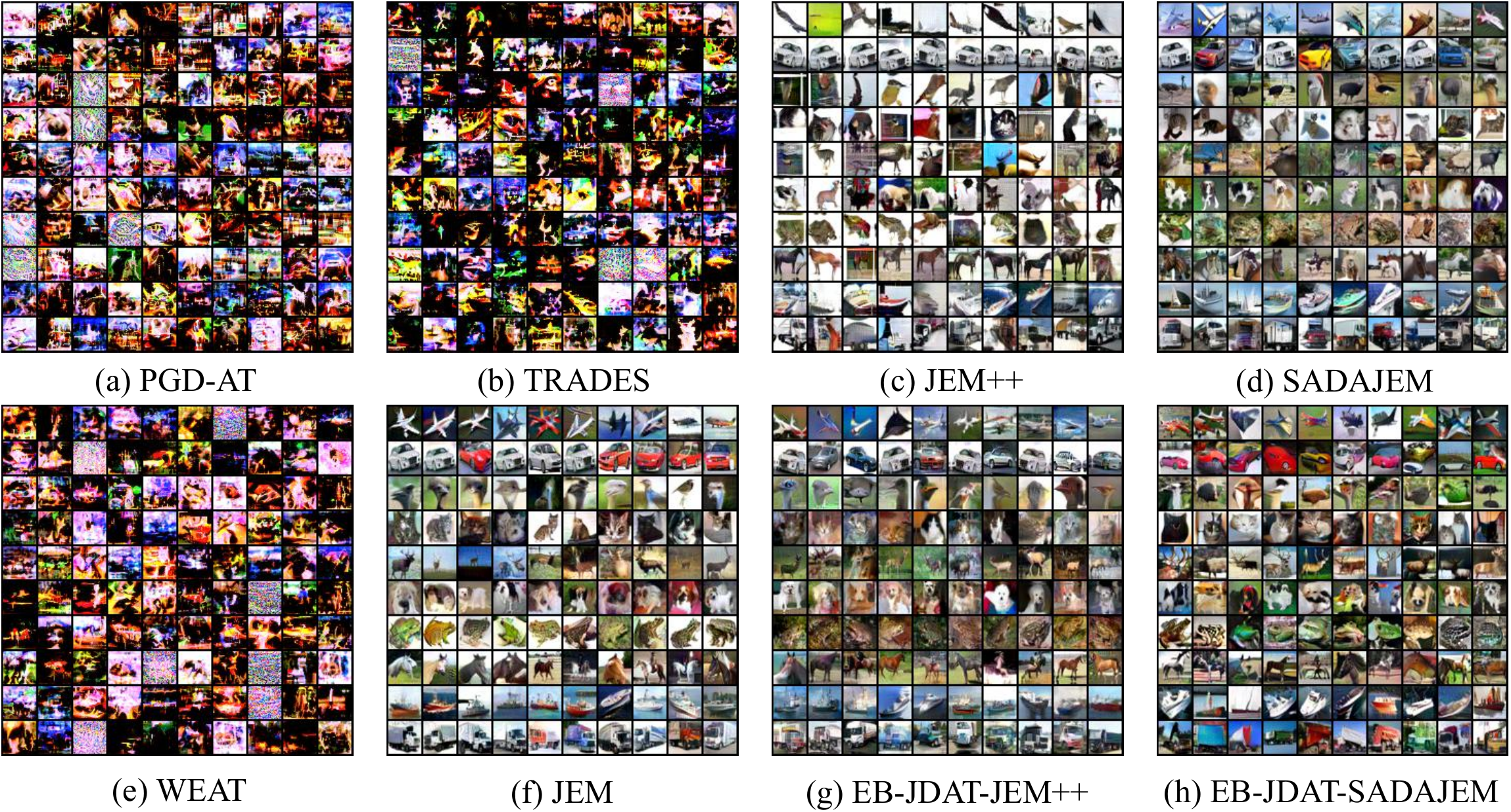} 
  \vspace{-0.1cm}
  \caption{Performance comparison of generated samples from different methods, all methods are sampled by SGLD with informative initialization.}
  \label{fig:generate_compare}
  \vspace{-0.5cm}
\end{figure*}

\subsection{Comparison Experiments And Analysis}
\paragraph{Compared with AT Methods}
We compare our approach with a range of AT methods, including three popular ones: MART \cite{wang2019improving}, AWP \cite{wu2020adversarial}, and LBGAT \cite{cui2021learnable}, as well as six SOTA methods: LAS-AWP \cite{jia2022adversarial}, UIAT \cite{dong2023enemy}, SGLR \cite{li2024soften}, TDAT \cite{tong2024taxonomy}, AGR-TRADES \cite{tong2024balancing}, and DHAT-CFA \cite{zhang2025towards}. For evaluation, we select the PGD-20 \cite{mkadry2017towards} and \KC{standard} AutoAttack (AA) \cite{croce2020reliable} which consists of APGD-CE \cite{croce2020reliable}, APGD-DLR \cite{croce2020reliable}, FAB \cite{croce2020minimally} and Square \cite{andriushchenko2020square}. Both attacks utilizing the $\ell_\infty$ norm and \KC{without any defense specific adjustments}. All ATs are training on WRN28-10 \cite{zagoruyko2016wide}. The results on CIFAR-10 and CIFAR-100 are shown in \cref{table:robustness_comparison}, where EB-JDAT not only maintains the classification accuracy of clean data (90.39\% on CIFAR-10; 68.35\% on CIFAR-100) but also achieves robustness that even surpasses existing SOTA AT methods (+6.38\% on CIFAR-10; +1.23\% on CIFAR-100). To evaluate the performance of our method on large-scale datasets, we further conducted comparative experiments on the ImageNet subset \cite{deng2009imagenet} in \cref{table:imagenet_subset_acc_robust}. Given that adversarial training on the full ImageNet dataset requires substantial computational resources \KC{(e.g., \textbf{38 hours on 128 x V100s}\cite{xie2019feature})}, we selected a subset consisting of the first 100 categories from ImageNet-1000 and down-sampled the training images to a resolution of $64 \times 64$. Thus, our method effectively balances the trade-off between accuracy on clean data and robustness against adversarial attacks. Furthermore, EB-JDAT is compatible with both SADAJEM \cite{yang2023towards} and JEM++ \cite{yang2021jem++}, thereby demonstrating the universality of our approach. 

\begin{figure}[t]
  \centering
  \includegraphics[width=\linewidth]{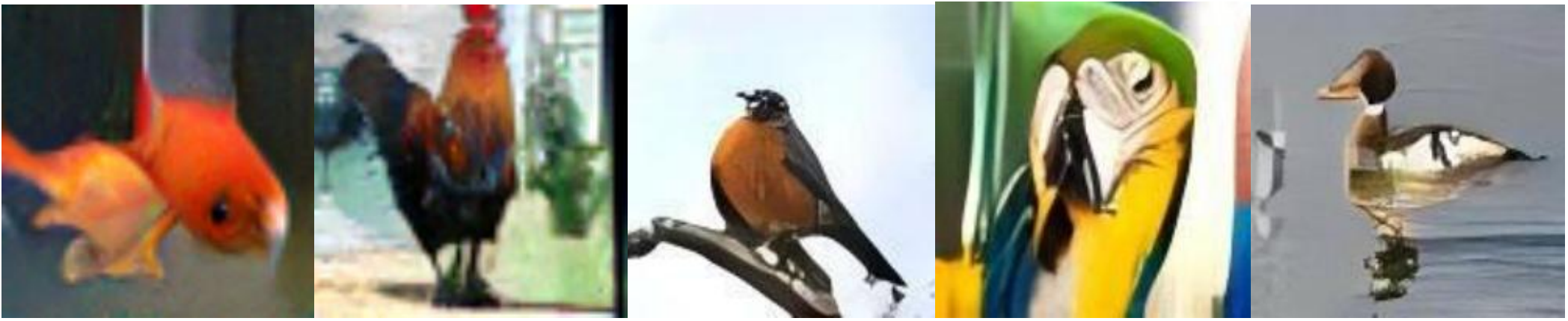} 
  \caption{Performance of generated samples on ImageNet subset (64x64) with EB-JDAT-JEM++.}
  \label{fig:gen_imagenet}
  \vspace{-0.6cm}
 
\end{figure}

\vspace{-0.3cm}
\paragraph{Comparisons with AT Using Generated Data.} Another strategy for improving robustness using generative models is to incorporate additional training data generated by generative models in AT~\cite{pang2022robustness, gowal2021improving}. Therefore we compare EB-JDAT with SCORE~\cite{pang2022robustness}, Better DM~\cite{wang2023better} and the method proposed in~\cite{gowal2021improving}. SOCRE and Better DM utilize the SOTA Diffusion models \cite{ho2020denoising} to generated addition training data, while \cite{gowal2021improving} generate addition data from a ensemble of DDPM \cite{ho2020denoising}, GANs \cite{goodfellow2014generative} and VDVAE \cite{childvery} generators. All methods are implemented using the WRN28-10 \cite{zagoruyko2016wide}. The results are presented in \cref{table:robustness_comparison_extradata}. \KC{For data-augmented AT, we estimate total training time by measuring the per-batch runtime (100 samples/batch) and extrapolating to the full schedule on one 3090 GPU.} Although EB-JDAT does not rely on additional generated data or increased training epochs, it still achieves competitive performance in terms of clean accuracy and robustness. EB-JDAT even outperforms SCORE-1M and \cite{gowal2021improving}. Unlike discriminative learning approaches \cite{pang2022robustness, gowal2021improving, wang2023better}, which struggles to capture the underlying structure of both clean and adversarial data distributions, EB-JDAT, as a hybrid generative model, provides more accurate gradient estimations of clean/adversarial data densities, leading to substantial improvements in both accuracy and robustness. Notably, this is achieved without requiring additional training data and with minimal computational overhead, \KC{compared with data-augmented AT}.

\vspace{-0.3cm}
\paragraph{Compared with JEMs and Energy-based AT.}
In this section, we compare EB-JDAT with other JEMs \cite{grathwohl2019your,yang2021jem++,yang2023towards} and energy-based AT \cite{zhu2021towards,mirza2024shedding} in terms of classification accuracy, robustness and generative performance. \cref{table:performance_comparison_gen} presents an evaluation of our method against JEM variants with respect to classification accuracy and generation quality. To assess generation quality, we employ the Fréchet Inception Distance (FID) and Inception Score (IS) as metrics. Compared to JEMs, EB-JDAT maintains comparable accuracy and generative performance, while EB-JDAT-SADAJEM even outperforms JEM (FID-10.98) and JEM++ (FID-9.7) in terms of generation. Moreover, EB-JDAT exhibits significantly improved robustness over JEMs, although it exhibits a slight decrease in classification accuracy and generative ability relative to the strongest JEMs \cite{yang2023m}, as shown in \cref{table:performance_comparison_gen}. This reflects an intrinsic trade-off: adversarial training uses adversarial examples to regularize decision boundaries and reshape the energy landscape, which can shift probability mass away from the clean data manifold and slightly reduce sample fidelity, which are shown in \cref{fig:robustness_vs_gen}. Overall, compared with existing JEMs \cite{grathwohl2019your,yang2021jem++,yang2023towards}, EB-JDAT achieves the best balance among classification, generation, and robustness.

Next, we compare EB-JDAT with energy-based AT \cite{zhu2021towards,mirza2024shedding}. First, we compare the accuracy and robustness with them, which is reported in \cref{fig:robustness_jem}. The results demonstrate that our method significantly enhances robustness with minimal accuracy degradation, outperforming existing energy-based AT. Moreover, our method yields superior generative performance compared to energy‐based methods, which are shown in \cref{table:performance_comparison_gen}. EB-JDAT surpasses JEAT and WEAT with an FID reduction of 10.82 and 3.32, respectively. Unlike JEAT \cite{zhu2021towards} directly introduces adversarial samples into the JEMs training process to model the joint distribution ${p}_{\theta}(\tilde{\mathbf{x}}, y)$, we model the joint probability $p_{\theta}(\mathbf{x}, \tilde{\mathbf{x}}, y)$ to capture the distribution of clean and adversarial samples more accurately, thus performing better in accuracy, robustness and generative ability. On top of that, WEAT \cite{mirza2024shedding} reinterprets TRADES \cite{zhang2019theoretically} as an EBM to model the conditional distribution $p_{\theta}(y \mid \tilde{\mathbf{x}}, \mathbf{x})$, which is still a discriminative model. Although \cite{mirza2024shedding} notes that PCA-based initialization can give WEAT generative capabilities (FID=30.74), but under the same comparison of informative initialization, WEAT shows poor generative performance (FID=177.92). Consequently, EB-JDAT outperforms previous energy-based AT in classification accuracy, robustness, and generative capability.

\vspace{-0.5cm}
\paragraph{Visual Quality Analysis.}
To further visually explore generative ability of EB-JDAT, we generate images via SGLD sampling with informative initialization for conventional AT-trained models, energy-based AT-models, JEMs, and EB-JDAT in \cref{fig:generate_compare}. As demonstrated in \cref{fig:generate_compare}, the conventional AT-trained models (a, b) and energy-based AT-models (e) possess limited generative capacity, resulting in difficulties in generating high-quality images. Conversely, EB-JDAT produce higher-resolution images with richer background details, which is more closely matching the features of the original dataset. In addition, EB-JDAT (g, h) surpasses JEM (f) and perform competitively with JEM++ and SADAJEM in generative capacity, producing more diverse and detailed images with only a slight reduction in sharpness, particularly in the 'car' category. We also implement our method on the ImageNet subset and generate higher resolution images, as shown in \cref{fig:gen_imagenet}.

\subsection{Ablation Study}
\label{Ablation Study}
\paragraph{Ablation of Weighted Gradient Components.}

In EB-JDAT, we employ a weighted combination of gradients, as detailed in Algorithm \ref{alg:EB-JDAT}, to compute the overall optimization gradient, expressed as $h_{\theta} = w_1 h_1 + w_2 h_2 + w_3 h_3$. To assess the individual contribution of each weight, we perform ablation studies by setting each of $w_1$, $w_2$, and $w_3$ to zero and evaluating their impact on overall performance, including (1) standard AT (NO.1) , (2) jointly modeling the energy distributions of adversarial and clean samples (NO.2) , and (3) jointly modeling the energy distributions of adversarial, clean, and generated samples (NO.3 and NO.4). The results, presented in Table \ref{table:ablation_performance_comparison}, reveal two key findings: (1) $\frac{\partial \log p_{\theta}(\tilde{{\mathbf{x}}} | {\mathbf{x}})}{\partial \theta}$ ($h_2$) is not only the key to bridge the gap of accuracy, generation and robustness, but also stabilizes the training phase, reducing the risk of model collapse. (2) $\frac{\partial \log p_{\theta}({\mathbf{x}})} {\partial\theta}$ ($h_1$) can improve the ability of model in both classification and generation. Consequently, experimental results indicate that $w_1 = 1$, $w_2 = 1$, and $w_3 = 1$ provides the \textbf{best trade-off} between accuracy, robustness, generative capability, and stability. This configuration is therefore adopted as the default setting. Additionally, reducing $w_2$ can improve classification accuracy and generative quality, albeit with a slight decrease in robustness.

\begin{figure}[t]
  \centering
  \includegraphics[width=0.8\linewidth]{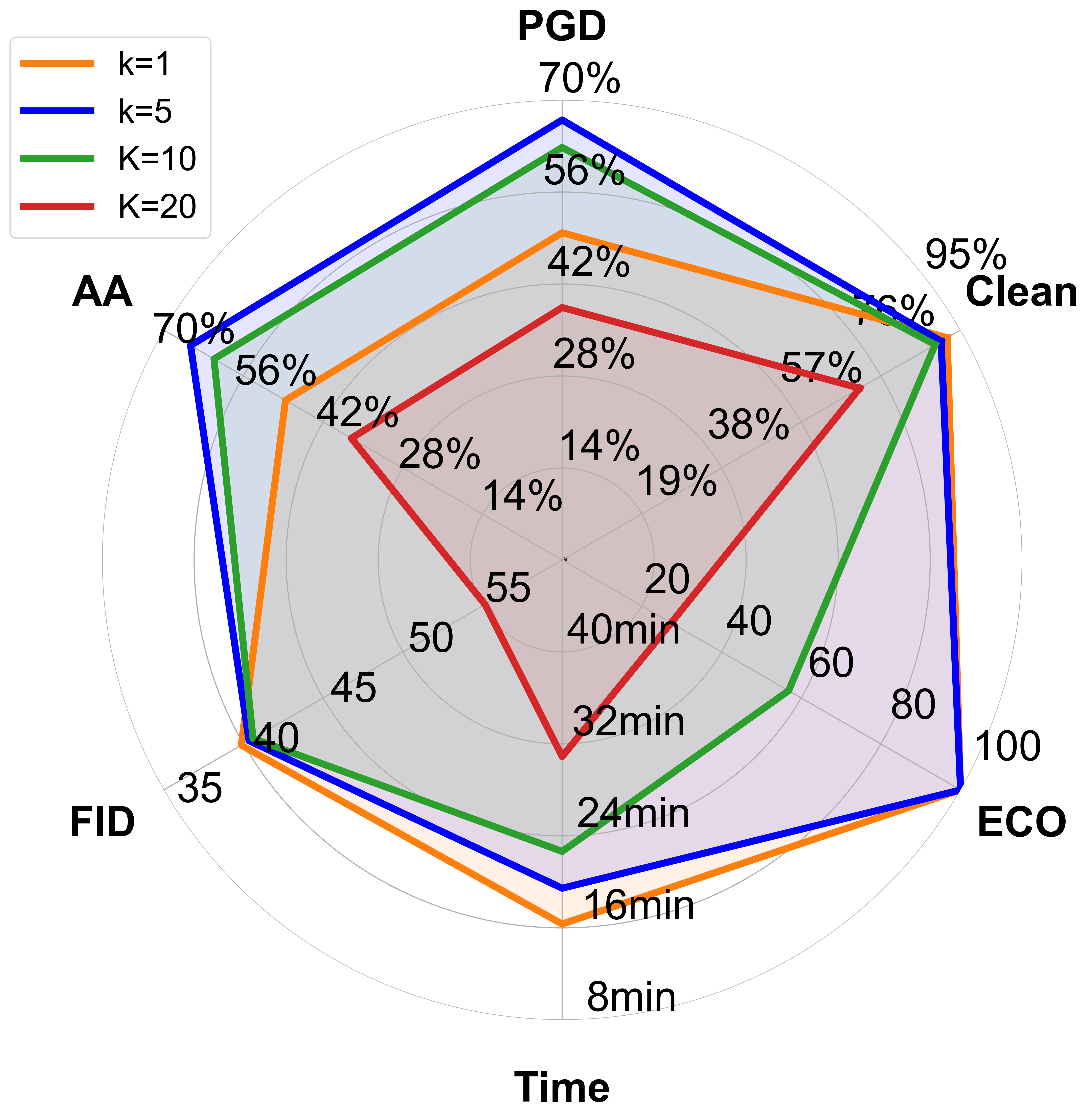} 
  \vspace{-0.3cm}
  \caption{Ablation study of adversarial sampling steps. ECO=100 indicates no collapse. Time denotes the per-epoch cost.}
  \label{fig:ablation}
  \vspace{-0.5cm}
\end{figure}

\vspace{-0.4cm}
\paragraph{Ablation of Adversarial Sampling Steps.}Further, we ablate the number of adversarial sampling steps K $\in$\{1,5,10,20\} while fixing the perturbation budget and step size. As shown in \cref{fig:ablation}, a setting of 5 steps provided the best trade-off, allowing the model to converge stably. This choice was made to balance two factors: finding sufficiently challenging adversarial examples and avoiding model collapse, which is a known issue for energy-based models on complex data, as we note in our limitations section .

\begin{table}[ht] 
\centering  
\caption{Ablation study of EB-JDAT-JEM++, ECO denotes the epoch at which collapse occurs. Results in bold indicate the best.}
\label{table:ablation_performance_comparison}
\small
\setlength{\tabcolsep}{4pt} 
\begin{tabular}{ccccccccccc}
\toprule
&NO&$w_1$ &$w_2$ &$w_3$& Clean (\%)$\uparrow$ & AA (\%)$\uparrow$ & FID$\downarrow$ & ECO \\
\midrule
&1&0 &0   &1 &88.95 &62.96 &173.53 &41 \\
&2&0 &1   &1 &89.84 &\textbf{64.69} &42.57 &n/a \\
&3&1 &0.5 &1 &\textbf{90.39} &64.09 &40.12 &n/a \\
\rowcolor[HTML]{D3D3D3}
&4&1 &1   &1 &90.37 &64.61 &\textbf{39.67} &n/a \\

\bottomrule
\end{tabular}
\end{table}
\vspace{-0.4cm}

\section{Conclusion}
In this work, we systematically investigate the underlying causes of the performance gap between discriminate classifiers, robust classifiers \cite{mkadry2017towards,zhang2019theoretically} and JEM-based classifiers \cite{grathwohl2019your,yang2023towards} in terms of accuracy, generation, and adversarial robustness, and propose a general and flexible optimization framework, called EB-JDAT to eliminate the inherent trilemma. Our experiments validate the effectiveness of these techniques across various benchmarks, demonstrating superior performance in achieving \KC{a new trade-off frontier ,} with particularly impressive SOTA results in robustness. 
\section{Acknowledgements}
This work was supported in part by the NSF China (No. 62302139, 62406320, 62472396), Grant of China Postdoctoral Science Foundation (NO.2025M781470) and Fundamental Research Funds for the Central Universities of China (PA2025IISL0113,JZ2025HGTB0227).
{
    \small
    \bibliographystyle{ieeenat_fullname}
    \bibliography{main}
}

\clearpage
\setcounter{page}{1}
\maketitlesupplementary

\section{Gradient obfuscation analysis}
\label{sec:rationale}
\cref{tab1} reports a per-attack AA breakdown, without using EOT, \cref{tab2} evaluates transfer attacks using PGD-AT as surrogate model, where EB-JDAT-JEM++ consistently outperforms a strong AT baseline. Following \cite{athalye2018obfuscated}, we check three masking indicators: (i) robustness not decreasing with larger distortion bound, (ii) one-step attacks outperforming iterative ones, and (iii) robustness relying on stochastic gradients. \cref{tab3} shows that robustness decreases with larger distortion bound, iterative attacks outperform one-step, and EOT-PGD is at least as strong as standard PGD, jointly contradicting gradient-masking indicators.

\begin{table}[t]
  \centering
  \small
  \caption{Robustness (\%) of EB-JDAT-JEM++ on CIFAR-10.}
  \scalebox{1}{
  \begin{tabular}{c|ccccc}
    \toprule
    Model  & APGD-CE& APGD-DLR &FAB&SQUARE \\
    \midrule
    Ours  & 64.46 & 65.29 & 90.52 & 70.72 \\
   
    \bottomrule
  \end{tabular}}
      \label{tab1}
\end{table}

\begin{table}[t]
  \centering
   \small
\caption{\setlength{\baselineskip}{1\baselineskip}Comparative robustness (\%) of EB-JDAT-JEM++ on CIFAR-10 under transfer attacks from a PGD-AT surrogate.
}

  \scalebox{1}{
  \begin{tabular}{c|ccccc}
    \toprule
    Model  & PGD& MI-FGSM &VMI-FGSM&VNI-FGSM \\
    \midrule
    TRADES  & 22.57 & 21.68 & 21.54 & 21.97 \\
    Ours  & \textbf{39.06} & \textbf{38.33} & \textbf{37.72} & \textbf{39.19} \\
   
    \bottomrule
  \end{tabular}}
      \label{tab2}
\end{table}

\begin{table}[t]
  \centering
  \small\caption{\setlength{\baselineskip}{1\baselineskip}Robustness (\%) of EB-JDAT-JEM++ on CIFAR-10 under APGD-CE (varying distortion/iterations) and EOT-PGD.}
  \setlength{\tabcolsep}{1.5pt}
  \scalebox{0.75}{
  \begin{tabular}{c|cccc|cccc|cccc}
    \toprule
    & \multicolumn{4}{c|}{APGD-CE: distortion bound} 
    & \multicolumn{4}{c|}{APGD-CE: iterations} 
    & \multicolumn{4}{c}{EOT-PGD: distortion bound} \\
    \cmidrule(lr){2-5}\cmidrule(lr){6-9}\cmidrule(lr){10-13}
    Model & 4/255 & 8/255 & 16/255 & 32/255 & 1 & 10 & 20 & 50 & 4/255 & 8/255 & 16/255 & 32/255 \\
    \midrule
    Ours  & 80.28 & 64.46 & 30.63 & 2.29  & 68.43 & 64.97 & 64.46 & 64.32 & 80.97 & 64.68 & 34.24 & 7.11 \\
    \bottomrule
  \end{tabular}}
  \label{tab3}
\end{table}

\section{Initialization for generation}
For fairness, we use the default initialization of each JEM variant, as mismatched training and evaluation initializations may lead to generation failure. JEM++/SADAJEM use informative init, while JEM (FID 38.40) uses random init, under which EB-JDAT-JEM achieves competitive generation {(FID 39.43)}.

\section{Limitation And Discussion}
\label{Limitation And Discussion}
Training EB-JDAT on complex, high-dimensional data remains challenging. This challenge is also encountered by JEMs,including JEM \cite{grathwohl2019your}, JEM++ \cite{yang2021jem++} and SADAJEM \cite{yang2023towards}. This instability arises from the typically sharp probability distribution of real data in high-dimensional space, which leads to inaccurate guidance for image sampling in regions with low data density. Although the approach we propose is a general and flexible optimization framework for all JEMs, considering training stability, we recommend training within faster and more stable JEM variants, such as JEM++ and SADAJEM. Nevertheless, our method significantly enhances the robustness of JEMs, surpassing SOTA AT \cite{zhang2025towards}, while incurring only a slight degradation in accuracy and generative performance, thereby achieving the \textbf{best overall trade-off} among robustness (68.76\%) , accuracy (90.39\%) , and generation (FID=27.42) . 

Scaling EB-JDAT to large-scale datasets remains challenging due to the high computational cost of adversarial training and the additional overhead of EBM sampling, we will clarify this limitation and explore latent-space optimization on larger datasets in future work.

\end{document}